\documentclass[runningheads]{llncs}
\usepackage{graphicx}
\usepackage[outdir=./]{epstopdf}
\usepackage{placeins}
\begin{document}
\title{Preprocessing and Modeling of Radial Fan Data for Health State Prediction}
\footnotetext[1]{The final publication is available at \url{https://link.springer.com/chapter/10.1007\%2F978-3-030-45093-9\_38}}

\author{Florian Holzinger \and Michael Kommenda}
\authorrunning{F. Holzinger and M. Kommenda}
\institute {
        Heuristic and Evolutionary Algorithms Laboratory\\
		University of Applied Sciences Upper Austria, Hagenberg\\
        \vspace{0.0cm}       
        \email{\{florian.holzinger,michael.kommenda\}@fh-hagenberg.at}\\
    }
    \maketitle

\begin{abstract}
Monitoring critical components of systems is a crucial step towards failure safety. Affordable sensors are available and the industry is in the process of introducing and extending monitoring solutions to improve product quality. Often, no expertise of how much data is required for a certain task (e.g. monitoring) exists. Especially in vital machinery, a trend to exaggerated sensors may be noticed, both in quality and in quantity. This often results in an excessive generation of data, which should be transferred, processed and stored nonetheless. In a previous case study, several sensors have been mounted on a healthy radial fan, which was later artificially damaged. The gathered data was used for modeling (and therefore monitoring) a healthy state. The models were evaluated on a dataset created by using a faulty impeller. This paper focuses on the reduction of this data through downsampling and binning. Different models are created with linear regression and random forest regression and the resulting difference in quality is discussed.

\end{abstract}

\section*{Introduction}
With the latest advance in electronics and computer science, many new technologies and trends have emerged. Sensors are becoming cheaper and more powerful, enabling the seamless integration of smartphones and wearables in our lives. Many people augment their life by carrying multiple sensors around, providing performance indications such as amount of steps taken today or average heart rate. The industry is changing in the same way, pushing several new trends such as Industry 4.0 and the Internet of Things. Machinery is being monitored, relevant data is collected, stored, and analysed. These trends induced a slow paradigm shift from preventive to predictive maintenance \cite{mobley2002introduction}, with the goal of predicting the current health state and the remaining useful lifetime of machinery. Imperative for any modeling approach or goal is the acquisition of relevant data. Radial fans \cite{Goetzler2015} are possible candidates for a predictive maintenance approach as they are often a crucial part of factories and an unforeseen outage may have a serious, negative impact. However, hardly any data from real world industrial radial fans is publicly available. 

Therefore, instead of pursuing the ambitious goal of calculating the remaining useful lifetime, this paper focuses on the necessary preprocessing steps and their influence on modeling the current health state of a specific radial fan. This radial fan is the key element of an ongoing project, exploring the applicability of predictive maintenance on industrial radial fans \cite{holzinger2018sensor,strumpf2019}. For this purpose, a test setup has been prepared and several different, relevant sensors are mounted on a radial fan. As the most common failure or sign of wear of radial fans is the abrasion and caking of an impeller, two different impellers were provided. One of them is new and flawless, and the other one artificially pre-damaged to simulate long-term abrasive stress. Several test runs have been carried out and data has been collected for each of these impellers. One of the current flaws of the setup is the huge volume of data generated, which roughly corresponds to about 1 GB/hour for one radial fan. This amount of data is acceptable for offline analysis but could pose a difficulty for online processing, be it for either monitoring or prediction purposes. 

Two major concerns of a predictive maintenance approach are the prediction quality and the costs, which are necessary to implement such a solution. As higher priced sensors tend to have a higher possible sampling rate, a reduction of sampling rate may render higher priced sensors superfluous and reduces the total cost of the system. Reducing the sampling rate is equivalent to a reduction of the total amount of data and ultimately a loss of information. The data loss can be mitigated by introducing additional features with common statistical key figures such as mean, variance or range while simultaneously reducing the total amount of data. There will always be a trade-off between the amount of data available and quality, but the decrease of quality can be attenuated by extracting beneficial features and reducing the sampling rate nonetheless. To achieve this goal, two different strategies are examined. First, a simple downsampling is applied and the resulting change in quality is shown. Second, the data is binned into bins of certain time intervals and several features are calculated, representing the bins. These calculated features are used in the modeling approach. 

For measuring the quality of the models created for the downsampling and binning approach, two different modeling strategies are used. These are linear regression (LR) \cite{draper2014applied} and random forest regression (RF) \cite{breiman2001random}.
The rotational speed was chosen as the target variable, as it can be manually adjusted. Therefore, additional test runs can be carried out, which can be used for later validation. 

\clearpage
\section*{Setup}
The setup for this paper consists of a radial fan with two differently prepared impellers. Each of them has a diameter of 625 mm, weights about 38kg and has 12 impeller blades. They show different signs of wear, the first one being in perfect condition, the second one was artificially damaged to imitate worn blades (abrasion at the edges), simulating long-term abrasive stress. The radial fan is powered by a 37 kW two-pole electric motor. The engine shaft is connected to the engine by a coupling. An electric frequency converter allows the adjustment of the rotational speed up to 2960 rpm and a control flap mimics pressure loss of the system. This setup is commonly found in industrial applications and therefore a suitable candidate for experimentation.

For the data-acquisition, four gyroscopes (BMX055) provide acceleration and rotation signals in all three dimensions. They were mounted on different positions on the bearings and the case. An additional sensor (GS100102) measures the rotational speed. The gyroscope signals are sampled with 1000 Hz, the rotational speed with 100 Hz. The setup can be extended with additional sensors to measure more properties (vibration, temperature, pressure, humidity, etc.). As for the downsampling and binning approach this paper focuses on, properties such as temperature and humidity with a significant, inherent inertia are not taken into account.

\begin{figure}
	\centering
	\includegraphics[width=1\textwidth]{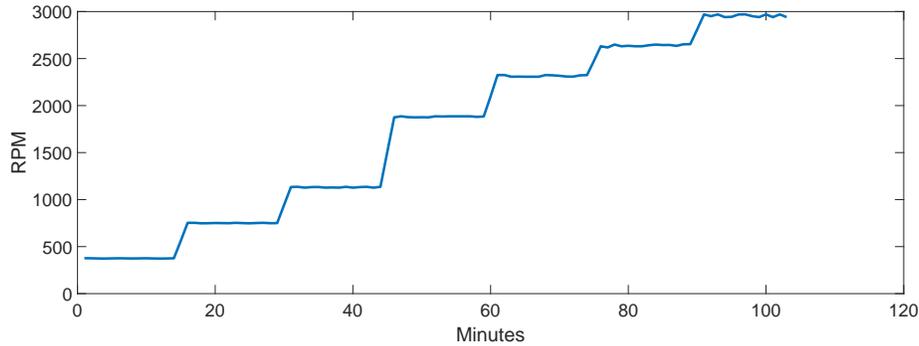}
	\caption[procedure]{Visualisation of a test run. The rotational speed was increased from 0 to 2960 rpm in 370 rpm-steps.}
	\label{fig:rpm_plot}
\end{figure}

A test routine was defined, enabling reproducible test runs and the generation of comparable data between different configurations (summer/winter, different impeller, etc.). To gather the data from different possible runtime configurations, the rotational speed was incremented stepwise from 0 rpm to 2960 rpm in 370 rpm-steps, each with a recording period of 15 minutes, resulting in a total observation period of about 2 hours. Figure~\ref{fig:rpm_plot} illustrates such a test run. Because the rotational speed is measured, slight noise is present in the raw data, which also increases with high rotational speed. 
This whole procedure was executed twice, differing in the mounted impeller and with otherwise, as far as possible, near identical parameters (same season, installation site, engine, control-flap position, etc.).

\section*{Methodology}
To summarize, two approaches to reduce the amount of data are examined. The first one is linear downsampling, the second one is binning the data into bins of certain time intervals and calculating features, representing the bins. 
Therefore, we chose six different configurations for the downsampling approach, using either 50 \%, 25 \%, 10 \%, 1 \%, 0.1 \%, or 0.01 \% of the available, raw data.
For the binning approach seven configurations were defined, using bin sizes of 100, 500, 1000, 2500, 5000, 10000 and 50000, containing consecutive samples of the raw data. Each of these bins is represented by key figures from descriptive statistics, namely the mean, standard deviation, range, median and kurtosis.
The available data, generated with the previously described setup was preprocessed (downsampled and binned, a total of 13 different configurations) and further split into two different training and test partitions for each of the runs. Firstly, the complete data was shuffled and the first 67 \% where used for training, the later 33 \% for test. 
Secondly, the training partition consists of the first, third, fifth and seventh rpm-step and the test partition was defined with the remaining three, as one of them was removed due to faulty data. The ascends in between each rpm-step were removed from both datasets. Also, considering the different sampling rates of the rotational speed and gyroscope sensors, the rotational speed was padded with the last measured value to remove missing values from the datasets. These datasets are later referred to as the shuffled dataset and the partitioned dataset. The shuffled dataset provides a general estimation of the change in quality with different downsampling rates and bin sizes, whereas the partitioned dataset suits for examining the ability of the chosen algorithms to generalize with the given configuration. The experiments were conducted with HeuristicLab, an open source framework for heuristic and evolutionary algorithms \cite{wagner2014}.

\section*{Results}
\begin{figure}
	\centering
	\includegraphics[width=1\textwidth]{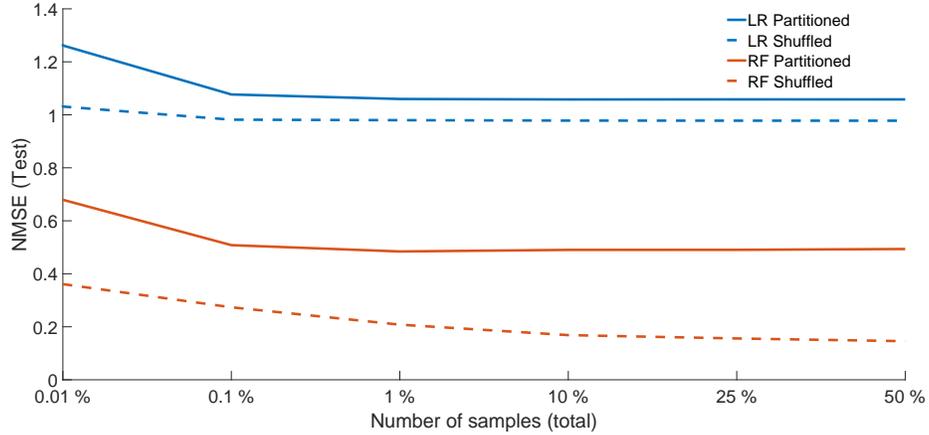}
	\caption[procedure]{Results of downsampling for LR and RF on the partitioned and shuffled dataset.}
	\label{fig:downsampling_both}
\end{figure}

As a baseline for the raw data, the NMSE of the shuffled dataset, using RF is 0.1398 (Test) and 0.0701 (Train) and the NMSE of the partitioned dataset is 0.4953 (Test) and 0.0419 (Train). For the LR, the corresponding values are 0.9774, 0.9773, 1.0586 and 0.9739.
The results of the previously defined setup and methodology for the downsampled datasets (both shuffled and partitioned) are illustrated in Figure~\ref{fig:downsampling_both}. In comparison with the results of the LR using raw models, only a marginal decrease in quality occurred. The more the data is reduced, the bigger the divergence of the NMSE between training and test gets.
For example, the NMSE for the LR on the shuffled dataset using 0.01 \% of the data is 1.0312 (Test) and 0.9048 (Train) and with the  partitioned dataset 1.2617 and 0.8589 respectively. The RF behaves similar on both partitions (e.g.  0.6792 (Test) and 0.1661 (Train) on the partitioned dataset using 0.01 \% of the data).
By considering the different character of the partitioned dataset, challenging the ability of the chosen algorithms to interpolate, the difference can be explained. For the given data, linear downsampling is a feasible way to reduce the total amount of data with little negative impact on the NMSE for both LR and RF with the best results being achieved by the RF.

\begin{figure}
	\centering
	\includegraphics[width=1\textwidth]{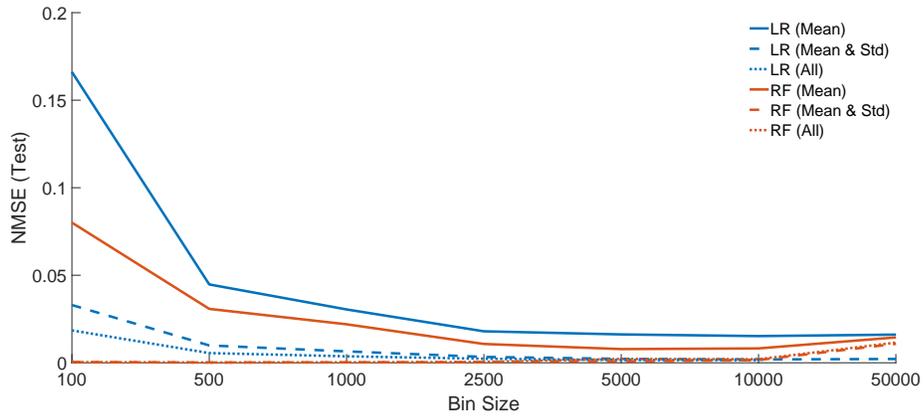}
	\caption[procedure]{Results of binning for LR and RF on the shuffled dataset.}
	\label{fig:binning_shuffled}
\end{figure}

\begin{figure}
	\centering
	\includegraphics[width=1\textwidth]{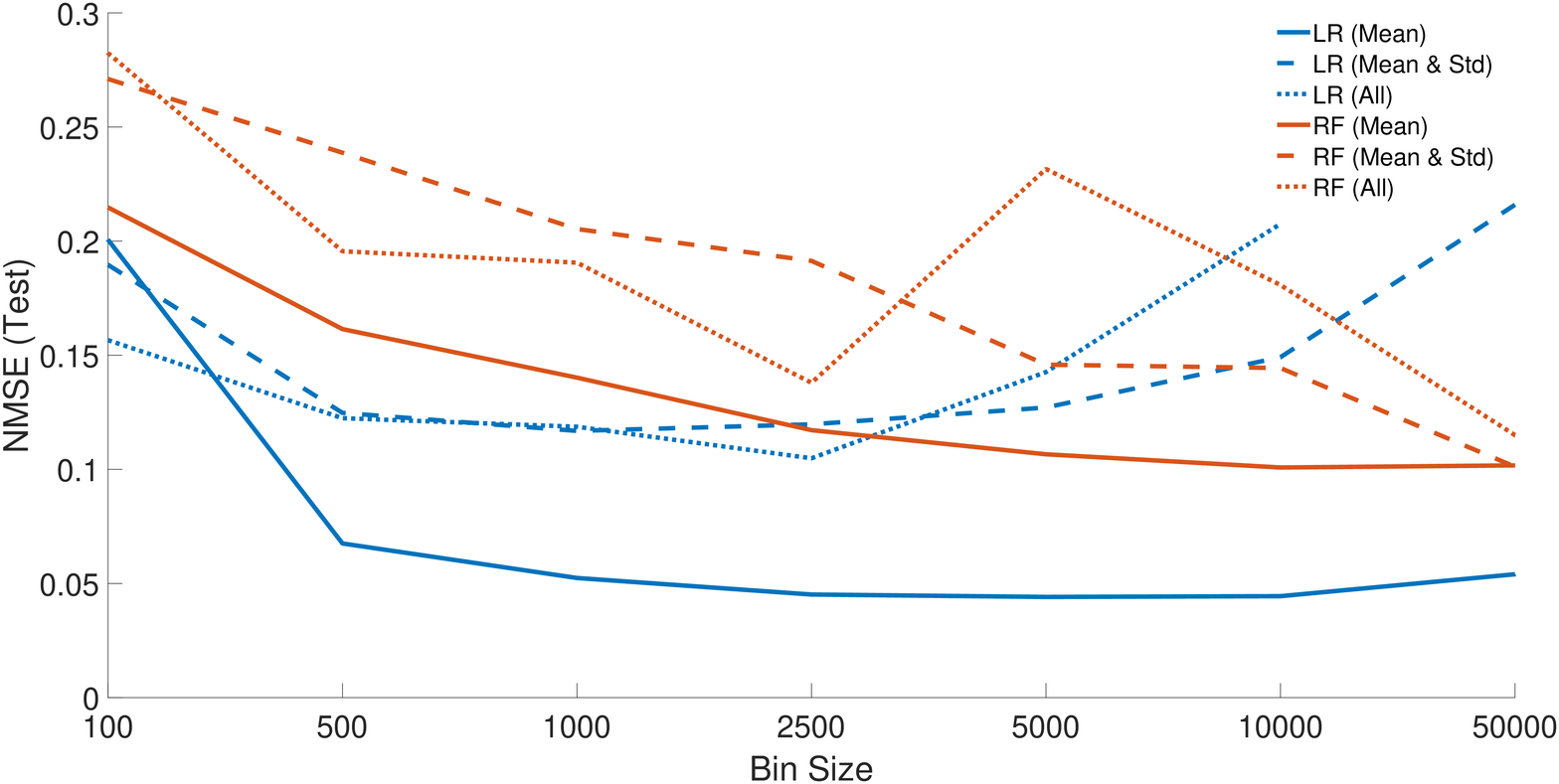}
	\caption[procedure]{Results of binning for LR and RF on the partitioned dataset.}
	\label{fig:binning_partitioned}
\end{figure}

Figure~\ref{fig:binning_shuffled} illustrates the results of the binning approach for the shuffled dataset, and Figure~\ref{fig:binning_partitioned} for the partitioned dataset. The algorithms are additionally configured to use either just the mean, the mean and the standard deviation, or all of the previously defined key figures.
In contrast to the downsampling approach, the NMSE improves with increasing bin size and yields better results. This seems reasonable, considering an existent noise, which is reduced by calculating the mean of a bin. By adding more features, the results improve when using the shuffled dataset and the RF creates better results than the LR again.
Just as with the downsampling approach, the partitioned dataset poses the bigger challenge for both algorithms. The best configuration was achieved by using LR with just the mean, as the RF apparently tends to overfit on the datasets, especially if more calculated features are added (NMSE of 0.1379 (Test) and 0.0003 (Train) for RF on the partitioned dataset with a bin size of 2500, using all available features, and 0.0004 and 0.0002 for the shuffled dataset, respectively).

\begin{figure}
	\centering
	\includegraphics[width=1\textwidth]{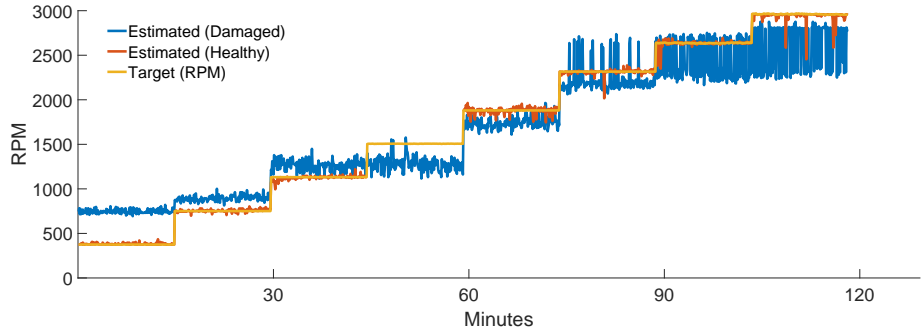}
	\caption[procedure]{Evaluation of a random forest model on the dataset with the healthy and damaged impeller.}
	\label{fig:eval_rf}
\end{figure}

As the goal is to reduce the amount of data while keeping an eye on the decline of model quality, which is possible according to the results, a differentiation between a healthy and damaged state (in our case discriminated by the impeller) must still be possible. Therefore, the models generated by the RF were evaluated on the corresponding dataset created with the faulty impeller.
The results of such an evaluation, using RF with a bin size of 5000 and utilizing all generated bin representatives on the shuffled dataset, are shown in Figure~\ref{fig:eval_rf}. The NMSE (calculated for the whole dataset) worsened from 0.0014 (healthy) to 0.0992 (damaged) and a visual distinction is possible. A similar behaviour can be observed for the other configurations as well, especially noteworthy for the LR using only the mean, which yielded the best results on the partitioned dataset (change of NMSE from 0.0146 (healthy) to 3.8796 (damaged), calculated for the whole dataset).

\FloatBarrier
\section*{Conclusion}
We analysed the impact of downsampling and binning on the quality of models generated with LR and RF to approximate the rotational speed of a radial fan with a healthy impeller, using data from four gyroscopes. The reduction of data achieved by using the downsampling approach had little influence on the model quality (NMSE) of both algorithms with the RF having the better model quality, up to a point where only 0.01 \% of the available data was used and a decline started to manifest. Similar behaviour could be observed by using the binning approach, even having a positive impact on model quality. The ability of the LR and RF to create generalizable models suffered from too many features, representing a bin. Using only the mean as a bin representation yielded the best results in this respect. Evaluating a selected model on an additional dataset generated with a damaged impeller illustrated that the so generated models can still be used to differ between a healthy and damaged state. By representing bins with a mean, the data is implicitly smoothed and possible noise is reduced. Binning may be used to reduce the data while simultaneously improving the model quality if the data has certain characteristics such as a stationary trend and existent noise. 

\section*{Acknowledgments}
The work described in this paper was done within the project \#862018 ``Predictive Maintenance für Industrie-Radialventilatoren'' funded by the Austrian Research Promotion Agency (FFG) and the Government of Upper Austria.
\bibliographystyle{splncs04}
\bibliography{EuroCAST2019_holzinger}
\end{document}